# Spinal nerve segmentation method and dataset construction in endoscopic surgical scenarios

Shaowu Peng[1(✉)], Pengcheng Zhao[1], Yongyu Ye[2], Junying Chen[1], Yunbing Chang[2], Xiaoqing Zheng[2]

[1] School of Software Engineering, South China University of Technology, Guangzhou, China
**swpeng@scut.edu.cn**
[2] Guangdong Provincial People's Hospital, Guangzhou, China

**Abstract.** Endoscopic surgery is currently an important treatment method in the field of spinal surgery and avoiding damage to the spinal nerves through video guidance is a key challenge. This paper presents the first real-time segmentation method for spinal nerves in endoscopic surgery, which provides crucial navigational information for surgeons. A finely annotated segmentation dataset of approximately 10,000 consecutive frames recorded during surgery is constructed for the first time for this field, addressing the problem of semantic segmentation. Based on this dataset, we propose FUnet (Frame-Unet), which achieves state-of-the-art performance by utilizing inter-frame information and self-attention mechanisms. We also conduct extended experiments on a similar polyp endoscopy video dataset and show that the model has good generalization ability with advantageous performance. The dataset and code of this work are presented at: https://github.com/zzzzzzpc/FUnet

**Keywords:** Video endoscopic spinal nerve segmentation, Self-attention, Inter-frame information.

## 1 Introduction

Spinal nerves play a crucial role in the body's sensory, motor, autonomic, and other physiological functions. Injuries to these nerves carry an extremely high risk and may even lead to paralysis. Minimally invasive endoscopic surgery is a common treatment option for spinal conditions, with great care taken to avoid damage to spinal nerves. However, the safety of such surgeries still heavily relies on the experience of the doctors, and there is an urgent need for computers to provide effective auxiliary information, such as real-time neural labeling and guidance in videos. Ongoing studies using deep learning methods to locate spinal nerves in endoscopic videos can be classified into two categories based on the level of visual granularity: coarse-grained and fine-grained tasks.

Coarse-grained vision task focuses on object detection of spinal nerve locations. In this task, object detection models applied to natural images are widely transferred to endoscopic spinal nerve images. Peng Cui et al. [1] used the Yolov3 [2] model to transfer training to the recognition of spinal nerves under endoscopy, which has attracted widespread attention. Sue Min Cho et al. [3] referred to Kai ming's work and achieved certain results using RetinaNet [4] for instrument recognition under spinal neuro-endoscopic images.

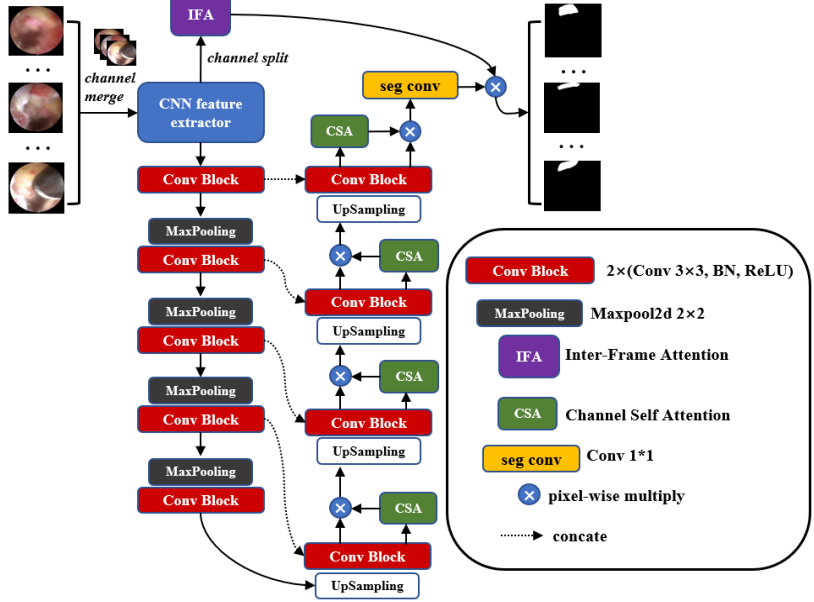

**Fig. 1.** Pipeline of the proposed FUnet, including the inter-frame attention module and channel self-attention module with global information.

Fine-grained tasks require semantic segmentation of spinal nerves, which provides finer contours and better visualization of the position and morphology of nerves in endoscopic view, leading to greater clinical significance and surgical auxiliary value. However, there are still very few deep learning-based studies on fine-grained segmentation of nerves, one important reason is a lack of semantic segmentation models suitable for spinal nerve endoscopy scenarios. The endoscopic image has the characteristics of blur, blisters, and the lens movement angle is not large, which is quite different from the natural scene image. Therefore, a segmentation model that performs well under natural images may not still be applicable under endoscopic images. Another reason is medical data involves ethical issues such as medical privacy, and the labeling of pixel-level data also relies on professional doctors. Furthermore, endoscopic images of the inside of the spine are often only available during surgery, which is much more difficult than obtaining image datasets from ordinary medical examinations. These lead to the scarcity of labeled data, and ultimately it is difficult to drive the training of neural network models.

In response to the above two problems, our contribution is as follows:

- We innovatively propose inter-frame and channel attention modules for the spinal neural segmentation problem. These two modules can be readily inserted into popular traditional segmentation networks such as Unet [5], resulting in a segmentation network proposed in this paper, called Frame-Unet (FUnet, Fig. 1). The purpose of the inter-frame attention module is to capture the highly similar context between certain adjacent frames, which are characterized by the slower movement of the lens and high similarity of information such as background, elastic motion, and texture between frames. Moreover, we devised a channel self-attention module with global information to overcome the loss of long-distance dependent information in traditional convolutional neural networks. FUnet achieved leading results in many indicators of the dataset we created. Furthermore, FUnet was verified on similar endoscopic video datasets (such as polyps), and the results demonstrate that it outperforms others, confirming our model's strong generalization performance instead of overfitting to a single dataset.
- We propose the first dataset on endoscopic spinal nerve segmentation from endoscopic surgeries, and each frame is finely labeled by professional labelers. The annotated results are also rechecked by professional surgeons.

## 2 Method

### 2.1 Details of Dataset

The dataset was taken by the professional SPINENDOS, SP081375.030 machines, and we collected nearly 10000 consecutive frames of video images, each with a resolution of up to 1080*1080 (Fig. 2). The dataset aims to address the specific task of avoiding nerves during spinal endoscopic surgery. To this end, we selected typical scenes from the authors' surgical videos that not only have neural tissue as the target but also reflect the visual and motion characteristics of such scenes. Unlike other similar datasets with interval frame labeling [6, 7], we labeled each image frame by frame, whether it contains spinal nerves or not. Each frame containing spinal nerves was outlined in detail as a semantic segmentation task. Additionally, we used images containing spinal nerves (approximately 4-5k images) to split the dataset into training, validation, and test sets for model construction (65%: 17.5%: 17.5%).

### 2.2 Network Architecture

The overall network architecture is based on a classical Unet network [5] (Fig. 1). In the input part of the network, we integrate $T$ frames ($T > 1$) in a batch to fully utilize and cooperate with the **inter-frame attention module** (***IFA***). The input features dimensions are ($B, T, C, H, W$), where $B$ is batchsize, $C$ denotes number of channels, $H$ and $W$ are the height and width of the image. Since we use a convolutional neural network for feature extraction, the 5-dimensional feature map is first merged into the convolutional network with $B$ and $T$ channels. Afterwards, the convolutional features are fed into the ***IFA*** for inter-frame information integration.

Unet's skip-connection method is a good complement to the information lost in the down-sampling reduction process, but it is difficult to capture the global contextual information due to the local dependency of the convolutional network. However, this global information is also crucial to the spinal nerve segmentation problem in the endoscopic scenario, so we inserted a **channel self-attention module (*CSA*)** in each up-sampling section with the global capability of the self-attention mechanism.

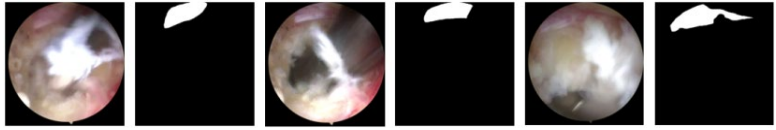

**Fig. 2.** Illustration of the original and labeled images.

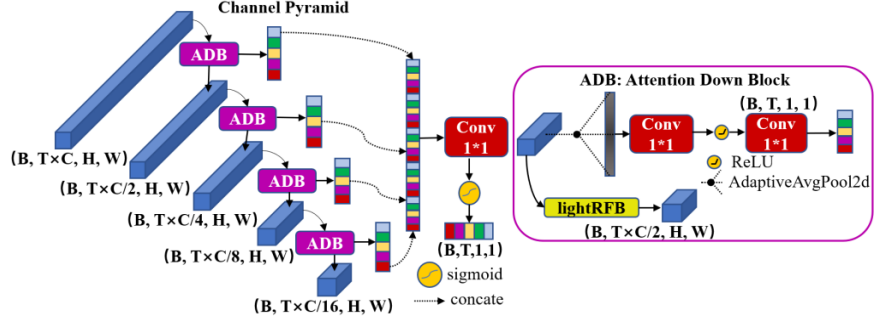

**Fig. 3.** Pipeline of inter-frame attention Module (IFA)

### 2.3 *IFA* Module

To exploit the rich semantic information (such as blur, blisters, and other background semantics) between endoscopic frames, we designed the *IFA* to correlate the weight assignment between T frames. (Fig. 3). If the features extracted by convolutional feature extraction are $(B \times T, C, H, W)$, we first need to split the features to obtain $(B, C \times T, H, W)$, which expands the channel dimension and allows subsequent convolutional operations to share the information between frames. After that, the feature matrix is down-sampled by four *Attention Down Block*'s (*ADB*) channel pyramid architecture to obtain the $(B, T, 1, 1)$ vector, and each $(1, 1)$ weight value of this vector in $T$ dimension will be assigned to the attention weight of the segmentation result of T frames.

**Channel Pyramid.** Although the multiscale operation in traditional convolutional neural networks can improve the generalization ability of the model to targets of different sizes [8, 9], it is difficult to capture information across frames, and down-sampling losses spatial precision at each cross-frame scale. Meanwhile, in many cases of endoscopic video, only the keyframes are clear enough to localize and segment the target, which can guide other blurred frames.

Hence, we propose a channel pyramid architecture to compress the channel dimension for capturing the cross-frame multiscale information, as well as to keep the feature map size unchanged in dimensions of height and width for preserving spatial precision. Such channel down-sampling obtains multi-scale information and semantic information in different cross-frame ranges. The result can adjust the frame weight on keyframe segmenting guidance. For detail, the feature matrix obtained by each ADB is compressed in the channel dimension, which avoids the loss of image size information. Like the perceptual field in the multi-scale approach, the number of inter-frame channels in the channel pyramid at different scales represents the magnitude of the scale across frames, and this inter-frame information at different scales is concated for further fusion calculations, which is used to generate attention weights.

**Attention Down Block (ADB).** This module (Fig. 3) is responsible for downsampling the channel dimension and generating the weight information between frames at one scale. We use the Light-RFB [10, 11] module for channel down-sampling without changing the size in the height and width dimensions. In terms of the generation of attention weights, the feature vector after the adaptive global average pooling operation will be scaled to the T dimension by two 1×1 convolutions.

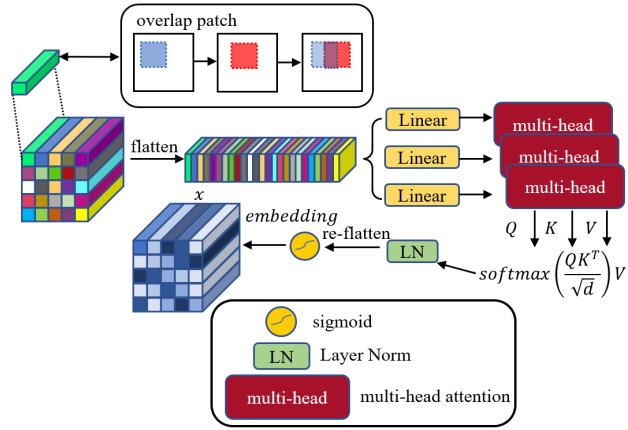

**Fig. 4.** Pipeline of Channel Self-Attention Module

## 2.4 *CSA* Module

Inspired by the related work of vision transformer [12, 13], we propose *CSA* mechanism (Fig. 4) to address the problem of the lack of global information of traditional convolutional networks under spinal nerve endoscopy videos. Different from the classical vision transformer work, firstly, in the image patching stage, we use the convolution operation to obtain a feature matrix with a smaller dimension (such as $B \times C \times 32 \times 32$), and the length corresponding to each patch is the length of each pixel channel ($32 \times 32$), which reduces the amount of computation while sharing of information between different patches. This is because, in the process of convolution down-sampling, the convolution kernels will naturally cause overlapping calculations on the feature map, which will lead to the increase of the receptive field and the overlap of information between different patches. We use three fully connected layers $Lq$, $Lk$, $Lv$ to generate Query, Key and Value feature vectors, this can be expressed as follows:

$$Query = Lq(X), \ Key = Lk(X), \ Value = Lv(X) \tag{1}$$

$X \in \mathbb{R}^{B \times (H \times W) \times C}$ is the original feature matrix expanded by the flatten operation. At the same time, to supplement the loss of spatial position information, we supplement the pos embedding vector by addition operation.

The multi-headed attention mechanism is implemented by a split operation. For calculation of self-attention, we use the common method [14], the Query matrix and the transpose matrix of Key are multiplied and then divided by the length $d$ of the embedding vector, and this part of the result will be multiplied with Value after soft-max operation. Finally, this part of the features operated by self-attention will enter the LN layer, and after the sigmoid operation, the dimensions of the final attention weight matrix are consistent with the input vector. The formula for this part is as follows:

$$\sigma \left\{ LN \left[ softmax \left( \frac{QK^T}{\sqrt{d}} \right) V \right] \right\} \tag{2}$$

## 3 Experiments

### 3.1 Implementation Details

**Dataset.** The self-built dataset is our first contribution. We set up the training set, test set, and validation set. For extended experiments on the polyp dataset, we used the same dataset configuration as PNS-Net [11]. The only difference is that in the testing phase, we used *Valid* and *Test* parts of the CVC-612 dataset [6] for testing (CVC-612-VT).

**Training.** On the self-built dataset and the CVC-612 dataset, we both use learning rate and weight decay of 1e-4 with Adam optimizer. On the self-built dataset, our model converges after about 30 epochs, while on the CVC-616 dataset, we use the same pre-training and fine-tuning approach as PNS-Net. Methods involved in the data augmentation phase include flipping and cropping. A single TITAN RTX 24GB graphics card is used for training.

**Testing.** Five images(*T=5*) are input to the *IFA* and use the same input resolution of 256*448 as PNSNet to ensure consistency in subsequent tests. Our FUnet is capable of inference at 75fps on a single TITAN RTX 24GB, which means that real-time endoscopic spinal nerve segmentation is possible in surgery.

### 3.2 Experimental Results

On our spinal nerve dataset, we tested the classical and leading medical segmentation networks Unet [5], Unet++ [15], TransUnet (ViT-Base) [16], SwinUnet (ViT-Base) [17] and PNSNet [11] respectively. For comparison, we used the default hyperparameter settings of the networks and employed a CNN feature extractor consistent with that of PNSNet.

Four metrics that are widely used in the field of medical image segmentation are chosen, maxIOU, maxDice, meanSpe/maxSpe and MAE. The quantitative result is in Table.1. Our FUnet achieves state of the art performance on different medical segmentation metrics. The Our-VT dataset means that we use the validation and test datasets for the testing phase (neither of which is involved in training).

The qualitative comparison is in Fig. 5, which shows our FUnet can more accurately segment the contour of the model and the texture of the edges.

**Table 1.** The quantitative results on our spinal nerve datasets. Our-VT means we use the valid and test part for testing. (Validation part is unseen during testing).

| Dataset | Model | Metrics | | | |
|---------|-------|---------|---------|---------|---------|
| | | maxDice ↑ | maxIOU ↑ | MAE ↓ | meanSpe ↑ |
| Our-VT | Unet | 0.778 | 0.702 | 0.044 | 0.881 |
| | Unet++ | 0.775 | 0.705 | 0.026 | 0.878 |
| | TransUnet | 0.825 | 0.758 | 0.082 | 0.881 |
| | SwinUnet | 0.823 | 0.752 | 0.026 | 0.877 |
| | PNSNet | 0.882 | 0.823 | 0.016 | 0.885 |
| | **FUnet** | **0.890** | **0.833** | **0.016** | **0.885** |

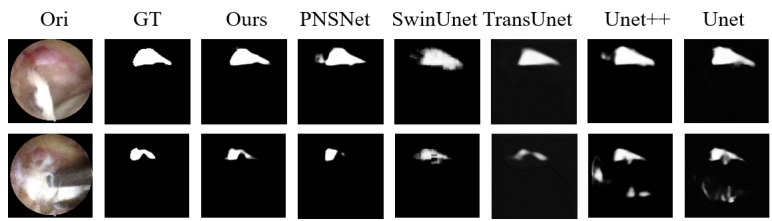

**Fig. 5.** The qualitative results on our dataset, for more results please refer to the supplemnetary material.

### 3.3 Ablation Experiments

The baseline model uses Res2Net [18] as the basic feature extractor with the Unet model, and in the first output feature layer, we adopt a feature fusion strategy consistent with PNSNet. We have gradually verified the performance of the *IFA* and *CSA* modules on the baseline model (Table. 2), and experiments have proved that our two modules can stably improve the segmentation performance of spinal nerves in endoscopic scene. A comparison of qualitative results is available in Fig. 6.

**Table 2.** Ablation studies. B* for baseline. I* for *IFA* module. C* for *CSA* module. Our-VT means we use the valid and test part for testing.

| Dataset | Model | Metrics | | | |
|---------|-------|---------|---------|---------|---------|
| | | maxDice ↑ | maxIOU ↑ | MAE ↓ | meanSpe ↑ |
| | B* | 0.862 | 0.798 | 0.016 | 0.885 |

| Our-VT | B*+I* | 0.884 | 0.826 | 0.015 | 0.884 |
| | B*+I*+C* | 0.890 | 0.833 | 0.016 | 0.885 |

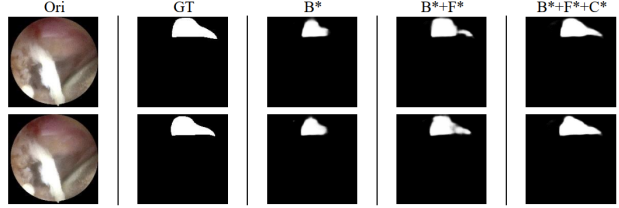

**Fig. 6.** The qualitative results on ablation study, for more results please refer to the supplemnetary material.

In addition, more extended experiments are carried out under a similar endoscopic polyp dataset CVC-612 [6] (Table. 3, CVC-612-TV means that we used both test and valid parts during the testing phase, the validation part was not visible during the training phase.), and the experiments show that our FUnet has good generalization performance and can adapt to endoscopic segmentation in different scenarios. A comparison of qualitative results is available in Fig. 7, our FUnet still performs well in its ability to segment the edges of polyps.

**Table 3.** The quantitative results on polyp datasets. CVC-612-TV means that we used both test and valid parts during the testing phase.

| Dataset | Model | Metrics | | | |
| | | maxDice ↑ | maxIOU ↑ | MAE ↓ | maxSpe ↑ |
| CVC-612-VT | Unet | 0.727 | 0.623 | 0.041 | 0.971 |
| | Unet++ | 0.713 | 0.603 | 0.042 | 0.963 |
| | PNSNet | 0.866 | 0.797 | 0.025 | **0.991** |
| | **FUnet** | **0.873** | **0.805** | **0.025** | 0.989 |

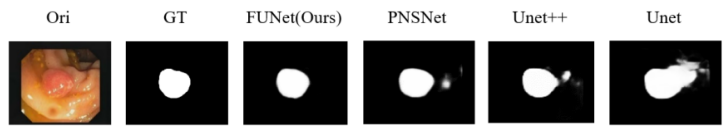

**Fig 7.** The qualitative results on CVC-612-Valid. for more results please refer to the supplemnetary material.

## 4 Conclusion

In this paper, we propose the industry's first semantic segmentation dataset of spinal nerves from endoscopic surgery to date and design the FUnet segmentation network based on inter-frame information and self-attention mechanism. FUnet has achieved state of the art performance on our dataset and shows strong generalization performance on polyp dataset with similar scenes. The *IFA* and *CSA* modules of FUnet can be easily incorporated into other networks. We plan to expand the dataset in the future with the help of self-supervised methods, to improve the performance of the model to provide better computer-assisted surgery capabilities for spinal endoscopy.